# Improving Deep Models of Person Re-identification for Cross-Dataset Usage

Sergey Rodionov[1,2], Alexey Potapov[1,3], Hugo Latapie[4], Enzo Fenoglio[4], Maxim Peterson[2,3]

[1] SingularityNET LLC
[2] Novamente LLC, USA
[3] ITMO University, Kronverkskiy pr. 49, 197101 St. Petersburg, Russia
[4] Chief Technology & Architecture Office, Cisco
{pas.aicv, astroseger, maxim.peterson}@gmail.com, {hlatapie, efenogli}@cisco.com

**Abstract.** Person re-identification (Re-ID) is the task of matching humans across cameras with non-overlapping views that has important applications in visual surveillance. Like other computer vision tasks, this task has gained much with the utilization of deep learning methods. However, existing solutions based on deep learning are usually trained and tested on samples taken from same datasets, while in practice one need to deploy Re-ID systems for new sets of cameras for which labeled data is unavailable. Here, we mitigate this problem for one state-of-the-art model, namely, metric embedding trained with the use of the triplet loss function, although our results can be extended to other models. The contribution of our work consists in developing a method of training the model on multiple datasets, and a method for its online practically unsupervised fine-tuning. These methods yield up to 19.1% improvement in Rank-1 score in the cross-dataset evaluation.

## 1 Introduction

Person tracking is one of the most typical tasks in visual surveillance. A great deal of methods for tracking within one camera has been developed. However, it is usually necessary to track a person using multiple cameras with non-overlapping fields of view. Here, traditional tracking techniques cannot be used, and the task of person re-identification (Re-ID) should be stated. This task is challenging because of high variations in background, illumination, viewpoint, human poses, etc., and absence of tight space-time constraints on candidate IDs like in tracking. Although some constraints do exist, and they should be used in a practical system.

Many attempts to solve this task exist [1], but it is far from being completely solved yet. Currently, deep convolutional neural networks (CNNs) are replacing traditional hand-crafted methods [2] that became possible due to both the progress in deep learning and the availability of larger public datasets like Market-1501 [3] and MARS [4], CUHK03 [5], DukeMTMC-reID [6], and others. Different deep learning models have been developed to solve the person Re-ID task including classification CNNs used for feature learning as in [7], Siamese CNNs that use image pairs [8], metric embedding CNNs trained with the triplet loss [9].

However, most of the existing deep learning models for Re-ID that show state-of-the-art results on different benchmarks are trained and tested on each of these benchmarks separately (e.g. [8, 9]). At the same time, in practice, it is usually necessary to deploy a person Re-ID system to a new camera set, for which a large labeled training set is expensive or impossible to acquire, so pre-trained models should be used. Unfortunately, as it is shown in [7], if a model is trained on one dataset and tested on another dataset, its performance drops significantly (seemingly below the level of hand-crafted features), because changes between datasets are rather large (see Figure 1). For example, Rank-1 score can decrease from 0.762 to 0.361 on the Market-1501 test set if the training was performed on the Duke training set instead of the Market-1501 training set.

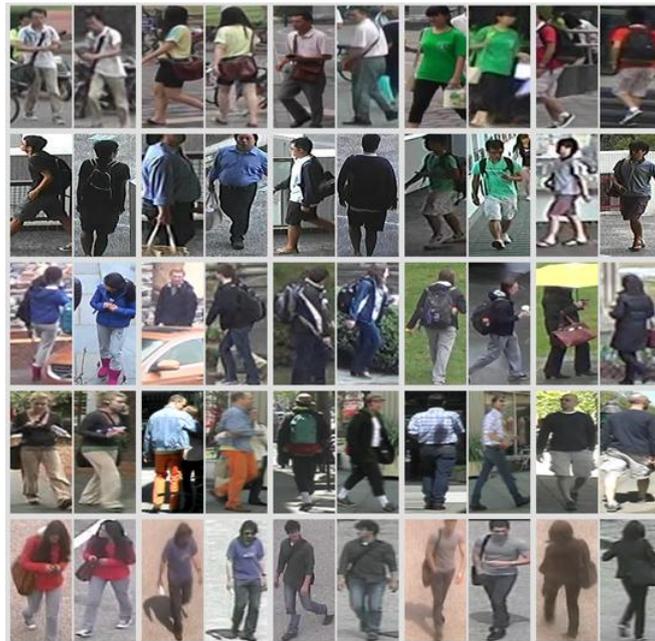

**Fig. 1.** Pairs of images of same IDs from different cameras from different datasets: Market-1501 [3], CUHK03 [5], Duke [6], Viper [14], WARD [15]

Few works address the problem of unsupervised fine-tuning of pre-trained models on new datasets, and even less of them report improvements achieved by unsupervised fine-tuning in comparison with corresponding pre-trained models, not just the final performance that depends both on the base model and unsupervised fine-tuning algorithm. One such recent work is [7], in which improvements in Rank-1 accuracy from 1.2% for the model trained on Duke and tested on CUHK03 to 11.9% for the model trained on CUHK03 and tested on Market-1501 are reported. While this can be useful in practice, the achieved accuracy is still far below supervised learning results. Also, the Progressive Unsupervised Learning (PUL) algorithm proposed in [7] assumes the known number of IDs in new dataset, which is usually unknown in practice. So, further progress in this task is needed.

In our work, we use one of the state-of-the-art models, namely, metric embedding network trained with the use of the triplet loss function [9] as the base model. The contribution of our paper is two-fold:
1) We develop the novel method to train the metric embedding network using the triplet loss function on multiple datasets resulting in the increased invariance (and corresponding cross-dataset scores) of the embedding w.r.t. cameras.
2) We develop a novel fine-tuning method for the person re-id task, which relies on easily accessible information to collect negative samples, so we call it practically unsupervised. It yields up to 10% improvement in Rank-1 score.

## 2 Metric Embedding Learning for Person Re-ID

**2.1. Loss Function**

In the task of person Re-ID, it is usually assumed that bounding boxes (BBs) around humans are already extracted. These BBs are usually resized to a certain fixed size. Each resized BB yields a pattern (image) in an initial space of raw features (color values in resized BBs) $\mathbf{x} \in R^F$.

BBs containing certain IDs can also be tracked within each camera forming the so-called tracklets, and in practice it is better to compare not separate BBs, but tracklets. But since it is usually enough to simply average over features calculated for BBs in a tracklet, and compare these averaged features, usage of tracklets doesn't influence on the rest functionality of a Re-ID system.

Each image $\mathbf{x}$ corresponds to a certain ID $y$, and the task is to identify which images from different cameras have same IDs. These IDs can be considered as classes, but it should be noted that the number of such classes is large and unknown while the number of images in each class is small. Thus, it is inefficient to cast the Re-ID task as the traditional pattern recognition problem.

One way to solve the task is to train a model (e.g. a Siamese Network) that accepts two images as input and infers if they correspond to same ID or not. The problem with this approach is that it requires running the model for one query image with each gallery image that is computationally expensive, especially in the case of deep neural networks.

Another possibility is to train a classification model (e.g. based on CNNs) for a fixed set of IDs known for a training set, and then to cut off the classification layer and compare images using high-level convolutional features, which were useful for classification. Similarity between images can be calculated directly as distance between these latent features, and the performance of such approach is acceptable in practice. However, images with the same ID will not be necessarily closer to each other than images with different IDs in the space of features useful for classification. Thus, additional step of metric learning is added to improve the overall performance.

In fact, what we want to learn is a metric embedding, i.e. a mapping $f(\mathbf{x}|\boldsymbol{\theta}):R^N \to R^M$ that transforms semantically similar images onto metrically close points in $R^M$ (and semantically dissimilar images onto metrically distant points). That is, $D_{i,j}=D(f(\mathbf{x}_i|\boldsymbol{\theta}), f(\mathbf{x}_j|\boldsymbol{\theta}))$ is small if $y_i=y_j$ and large otherwise, where $D$ is some distance measure (e.g. Euclidean). One can try to learn this mapping directly without learning surrogate classification model, if an appropriate loss function is specified.

The following triplet loss function can be used [9]

$$L_{tri}(\boldsymbol{\theta}) = \sum_{\substack{a,p,n \\ y_a=y_p \neq y_n}} [m + D_{a,p} - D_{a,n}]_+ , \qquad (1)$$

where $m$ is some margin by which positive and negative examples should be separated. That is, different triplets of images are considered – one is anchor image with index $a$, another one is a positive example $y_p=y_a$ with index $p$, and the last one is a negative example $y_n \neq y_a$ with index $n$, and we want distance $D_{a,p}$ be smaller than distance $D_{a,n}$ by $m$.

Softplus $\ln(1+\exp(x))$ is proposed to be used in place of the hidge function $[m+\bullet]_+$ in [9], since in Re-ID we want to pull images with the same ID further even after the margin $m$ is reached.

However, distances to similarly looking positives are easy to minimize and distances to arbitrary negatives are easy to maximize. Since triplets in (1) are mostly *weak* examples and all of them are also computationally demanding to use, *hard positive samples* and *hard negative samples* should be somehow selected to make embedding learning with the triplet loss successful. Computationally efficient selection of such hard samples can be done with the use of *Batch Hard* loss function [9]. The idea is to form batches using $P$ randomly selected classes (IDs) with randomly sampled $K$ images per class, and to select the hardest positive and negative samples within the batch to form the triplets for the loss function (see details in [9]).

Here, we also use soft margin and batch hard loss.

### 2.2. Model

We implemented the same network architecture as in [9] with few differences. Instead of ResNet-50, we used MobileNet [10], since we found that performance is very similar while MobileNet is much faster. We also discarded the last classification layer and added two fully connected layers to map high-level convolutional features to the embedding space. Similar to [9], we used the first dense layer with 1024 units with ReLU activation function [11], while the second (output) layer had 128 units corresponding to the embedding dimension. We also used batch normalization [12] between layers. Additionally we used dropout [13] after batch normalization, but in one experiment we found it beneficial to switch it off.

For the usual training, we used ADAM optimizer with default parameters (beta$_1$=0.9, beta$_2$=0.999). The learning rate was set to $10^{-4}$ during first 100 epochs. During next 300 epochs we exponentially decay the learning rate till $10^{-7}$. The number of steps per epoch was somewhat arbitrarily defined as $N_{total}/N_{batch}$, where $N_{total}$ is total the number of images in all used datasets, and $N_{batch}$=KP is the batch size. We used K=4 and P=18 in all experiments.

In our online fine-tuning method (described below), we used RMSProp optimizer with default parameters. The initial learning rate was set to $10^{-5}$, and the final learning rate after decay was $10^{-6}$. The number of steps per epoch was defined as $N_{pos}/K$, where $N_{pos}$ is total number of images in all positive samples. Since fine-tuning it was applied to the pre-trained model MARS is much larger, the number of epochs was taken smaller (20 epochs for Market-1501, and 2 epochs for MARS).

## 3 Improvements

### 3.1. Embedding Learning on Multiple Datasets

Usually the difference between different Re-ID datasets (for example between Duke and CUHK03) is quite significant. This is the source of the problem with using on one dataset a model trained on another dataset. The obvious idea is to train a model on several datasets simultaneously to force it to learn dataset-invariant embedding, which will be better transferrable to new datasets.

However, large difference between datasets means that it would be quite easy for the model to distinguish images from different sources. One can argue that it will prevent the model from learning invariant features. Instead, the model will focus on pushing different datasets apart in the embedding space, and on learning features more specific for individual datasets, which will be less useful in general setting.

In the case when we use the batch hard triplet loss, we can try to prevent this problem. Our approach is to train an embedding in such a way that network never "sees" images from different datasets simultaneously. We do it by forming each batch with images from only one dataset and we continuously switch between them during training. As we expect, this should prevent the model from simply pushing images from different datasets apart. Instead this should force it to search for invariant features, which, as we hope, will be useful for other datasets. We call this algorithm *BH-switch*. We will compare this algorithm with straightforward approach for training on multiple datasets where we just simply merge datasets together. We call this straightforward algorithm *BH-merge*.

### 3.2. Practically Unsupervised Fine-tuning

We can hope that the model trained on multiple datasets will be more dataset-invariant, and thus easier transferrable to new cameras. However, the model still needs to be tuned to achieve higher performance on new data distributions. This tuning should be done unsupervisingly. To apply the triplet loss directly, one should somehow guess positive and negative samples, i.e. which pairs of images correspond to same ID, and which correspond to different IDs.

### 3.2.1. Extracting Positive Samples

We propose the following method for extracting positive samples from unlabeled set of images (or tracklets) from two cameras. We can assume that we have a pre-trained model, which has a reasonable Re-ID performance for these two cameras. Using this model we calculate features for all images in our unlabeled training set. Then, the distance in the feature space is calculated between each possible pair of images from different cameras. We can except that pairs with minimal distance will be, with high probability, positive samples. We select first $N_p$ pairs with minimal distance and we will use them as presumable positive samples, where $N_p$ is a parameter.

Appropriate choice of $N_p$ can be important. If it is too small, there will be not enough training data to tune the model. If it is too high, the fraction of false positives among selected $N_p$ pairs will higher, and the model drift will take place. It is clear that $N_p$ should be proportional to the number of available images, but the coefficient of proportionality depends on two unknown factors, namely, on how well the pre-trained model suits these new data, and on the fraction of images (or tracklets) which belong to IDs which presented on both cameras (some IDs can be presented only on one camera). For example, the latter varies considerably for different pairs of cameras in MARS (from 0.24 to 0.92).

Our experiments showed that the dataset-dependent choice of $N_p$ influences the performance of fine-tuning. However, if we simply use $N_p = \alpha \min(N_1, N_2)$, where $N_{1,2}$ are the numbers of images (or tracklets) from two cameras, and $\alpha$ is a constant (we used $\alpha=0.1$), then the final performance drops no more than by 0.5%. Thus, we used this simple method without assuming the availability of additional information.

### 3.2.2. Extracting Negative Samples

Now we turn to the extraction of negative samples since we need them in the triplet loss. When the positive samples are defined, the negative sample can be extracted in a practically unsupervised manner. The simplest way is to take images, which were observed together with one of presumably positive images at the same time on the same camera. Even more negative samples can be gathered if two cameras have strictly non-overlapping field of view. In this case, we can use all images observed simultaneously on both cameras as negatives for each other. Thus, in the following experiments we will assume that we can obtain e.g. 10 negative samples for each positive pair.

### 3.2.3. Batch Hard Modification

Previously described Batch Hard loss should be adapted to the usage with our method for selecting positive and negative samples, since we don't have complete information about positive and negatives samples for each image in the batch.

We also form $P{\times}K$ batches, but each of them contains $K$ mutually positive samples (corresponding to one ID) and $(P{-}1)K$ images of other IDs, which are negative samples for first $K$ samples (but we don't know, what images have same IDs among these $(P{-}1)K$ images). While the original Batch Hard averages over all $P{\times}K$ images in the batch, here we average over first $K$ images, for each of which the hardest positive and negative samples are found. Consequently, this modified loss will be much noisier. In the following experiments with unsupervised fine-tuning, RMSProp optimizer (instead of ADAM which we used in Multiple datasets training) and a smaller learning rate appeared to be more efficient, since this modified loss is more difficult to optimize.

## 4  Experiments

**4.1. Multiple Datasets**

Let us start with testing the embedding training on multiple datasets with the following details. All the scores were computed without test-time data augmentation on MARS dataset. From the Duke dataset, we removed IDs, which belong to images only from a single camera. We found that such IDs only decrease the cross-camera Re-ID performance.

We underline that the models were trained on datasets which do not include MARS or Market-1501, but all scores are given for MARS dataset. So, we are focusing on the cross-base performance here. Table 1 shows Rank-1 and mAP scores for *BH-merge* and *BH-switch*.

**Table 1.** Rank-1 (mAP) scores on MARS for the embedding trained on different datasets

| Method<br>Training sets | *BH-merge* | *BH-switch* |
|---|---|---|
| **Duke** | 0.401 (0.204) | |
| **CUHK03** | 0.358 (0.189) | |
| **Duke + CUHK03** | 0.427 (0.243) | 0.455 (0.266) |
| **Duke + CUHK + WARD** | 0.437 (0.247) | 0.468 (0.269) |
| **Duke + CUHK + WARD + VIPER** | 0.444 (0.252) | 0.483 (0.296) |

As can be seen, training the embedding on multiple datasets without fine-tuning results in better performance even in the case of *BH-merge*. Even if the embedding shows an inferior performance on some dataset (e.g. CUHK03 in comparison to Duke), adding this dataset to the united training set improves the cross-base performance. Adding VIPER [14] and WARD [15] datasets also improved the performance. One can also see, that our *BH-switch* method is better then *BH-merge* (by almost 4%).

## 4.2. Unsupervised Fine-Tuning

We used the model trained on CUHK03+DUKE+WARD+VIPER as the pre-trained model for further fine-tuning. Its performance before fine-tuning was 0.483 / 0.296 (Rank-1 / mAP). The tests were conducted on MARS dataset, which contains tracklets from 6 cameras.

We extracted presumably positive samples for each combination of cameras (15 possible combinations in total). Each of our "presumably positive" sample is a pair of tracklets from different cameras. We expect that most of our "presumably positive" pairs are indeed belong to the same ID, but of course we will have some fraction of errors. For each positive sample we choose 10 negative samples (random tracklets from the selected pair of cameras, which do not belongs to the ID of the first tracklet from our "positive" pair), which are easily available in practical situations. One should note that if a "presumably positive" pair is identified mistakenly, there is some possibility that negative samples will also contain errors (some negative samples can belong to ID of the second tracklet from the "presumably positive" pair).

First of all, we checked the fraction of real positive samples among selected "presumably positive" samples. This fraction varies from 0.6 to 0.99 for different pairs of cameras. For most of them, it is larger than 0.8, and the average value is 0.83.

The Rank-1 / mAP scores after fine-tuning with our method appeared to be 0.566 / 0.355 meaning +8.3% improvement in Rank-1 over the model pre-trained on multiple datasets and +16.5% improvement over the best embedding model pre-trained on a single dataset. These results were obtained using the training set of MARS for fine-tuning. Since the goal is online unsupervised fine-tuning, both training and test sets are acceptable to use. So, we performed an additional step: we took the model fine-tuned on the training set and additionally fine-tuned it on the test set. The final scores increased to 0.592 / 0.380, so the overall improvement in Rank-1 is +19.1%.

Although our method is developed for real situations, in which tracklets are usually available, in many datasets they are absent, so we compare our results with [7] on Market-1501 dataset without tracklets. We also note that we didn't use dropout in our model here, because it leads to slightly worse results (unlike on MARS, because MARS contains incorrect labels). Here, we also used test-time data augmentation (see [9] for details). The comparison results are shown in Table 2.

**Table 2.** Rank-1 (mAP) scores on Market-1501 for the embedding and PUL trained on different datasets

| Method / Training sets | PUL (no fine-tuning) | PUL | Our (no fine-tuning) | Our |
|---|---|---|---|---|
| **CUHK03** | 0.300 (0.115) | 0.419 (0.180) | 0.337 (0.143) | 0.397 (0.183) |
| **Duke** | 0.361 (0.142) | 0.447 (0.201) | 0.417 (0.175) | 0.509 (0.238) |
| **Multiple datasets** | 0.400 (0.170) | 0.455 (0.205) | 0.518 (0.208) | 0.608 (0.350) |

It should be noted that the baseline model for PUL was pre-trained on Duke+CUHK03 in the case of "multiple datasets". The authors [7] point out "we find that initialization using more labeled datasets does not noticeably improve re-ID accuracy. Sometimes, multi-dataset initialization yields even worse results than single dataset initialization". Thus, we consider the pre-training of our model on more datasets as an improvement.

It should also be noted that we performed fine-tuning of our model on a united test and training sets of Market-1501, because no labels are used, and in a real situation one is interested in the online fine-tuning. If we use only the training set for fine-tuning, the scores are somewhat lower 0.586 (0.333).

Since we and [7] use different baseline models, not the absolute scores, but improvements are of interest. From Table 2 it can be seen that the embedding baseline model is better (and thus actually more difficult for improvements), but also relative increase in scores are higher in our method both for multiple dataset pre-training, and for online fine-tuning. Indeed, PUL gains only +0.8% in Rank-1 due to the use of more datasets (in addition to Duke) during pre-training, and +5.5% due to fine-tuning on several datasets (with overall +9.4% in Rank-1 when switching from Duke-baseline to Multiple datasets-PUL). Our fine-tuned model gains +9.9% thanks to the pre-training on several datasets and +9.0% due to fine-tuning on several datasets (with overall +19.1% in Rank-1), which is considerably larger improvement in comparison to PUL (although PUL has superior improvement in the case of CUHK03 only).

## 5 Conclusion

In this paper, we have proposed two practical techniques, which help to improve performance of Re-ID systems in real situations, in which these systems should be applied to unlabeled images taken by new cameras placed at new locations. We model this situation by the cross-dataset testing using different public datasets, and take the metric embedding learning with the triplet loss function as a baseline model.

The first technique consists in a special formation of batches using images from several datasets. The main idea is not to feed images from different datasets to the embedding network simultaneously, so the model will learn to distinguish different IDs from the same dataset, but not to distinguish datasets. As a result, we achieved +8.2% increase in Rank-1 score on MARS dataset due to the training on 4 datasets (Duke, CUHK03, WARD, VIPER) instead of using one best dataset (Duke). At the same time, simple merging of datasets yields only +4.3%.

The second technique consists in selecting presumably positive and negative samples to perform fine-tuning from the target unlabeled dataset, for which the Re-ID system should be applied. The main idea is to supplement best positive samples guessed by the model with samples, which appear with presumably positive pairs simultaneously in the video frames and which thus should be negative samples. Our fine-tuning algorithm results in +9% improvement in Rank-1 score on MARS and

Market datasets (with the assumption that we can obtain 10 negative examples for each presumably positive pair).

The overall improvement in Rank-1 on MARS and Market-1501 achieved by switching from the baseline embedding model trained on Duke to the fine-tuned model pre-trained on several datasets was +19.1%, which is much higher than +9.4% achieved by PUL [7].